\documentclass[preprint,1p]{elsarticle}
\usepackage{graphicx}
\usepackage{gensymb}
\usepackage{subfigure}
\usepackage{comment}
\usepackage{booktabs}
\usepackage{mathtools}
\usepackage{multirow}
\usepackage{threeparttable}
\usepackage{amsmath,amssymb} 
\usepackage{amsthm}
\usepackage{color}
\usepackage{algorithm}
\usepackage{algorithmic}
\usepackage{makecell}
\usepackage{bbding}
\usepackage{soul}
\usepackage{ulem}
\usepackage{bbm}
\usepackage{url}

\usepackage{amssymb}

\newtheorem{example}{Example}

\newtheorem{definition}{Definition}
\newtheorem{problem}{Problem}

\begin{document}

\begin{frontmatter}



\title{Pre-Training on Dynamic Graph Neural Networks} 


\author[1,2]{Ke-Jia CHEN\corref{cor1}}
\ead{chenkj@njupt.edu.cn}

\author[1]{Jiajun ZHANG}

\author[1]{Linpu JIANG}

\author[1,2]{Yunyun WANG}

\author[1]{Yuxuan Dai}

\cortext[cor1]{Corresponding author}

\address[1]{School of Computer Science, Nanjing University of Posts and Telecommunications, Nanjing, China }
\address[2]{Jiangsu Key Laboratory of Big Data Security \& Intelligent Processing, Nanjing University of Posts and Telecommunications, Nanjing, China}

\begin{abstract}
The pre-training of the graph neural network model is to learn the general characteristics of large-scale graphs or the similar type of graphs usually through a self-supervised method, which allows the model to work even when node labels are missing. However, the existing pre-training methods do not take the temporal information of edge generation and the  evolution process of graph into consideration. To address this issue, this paper proposes a pre-training method based on dynamic graph neural networks (PT-DGNN), which uses the dynamic graph generation task to simultaneously learn the structure, semantics, and evolution features of the graph. The method mainly includes two steps: 1) dynamic subgraph sampling, and 2) pre-training using two graph generation tasks. The former preserves the local time-aware structure of the original graph by sampling the latest and frequently interacting nodes. The latter uses observed edges to predict unobserved edges to capture the evolutionary characteristics of the network. Comparative experiments on three realistic dynamic network datasets show that the proposed pre-training method achieves the best results on the link prediction fine-tuning task and the ablation study and further verifies the effectiveness of the above two steps.

\end{abstract}


\end{frontmatter}



\section{Introduction}

In order to explore the learning ability of neural networks in structural data like graphs, graph neural networks (GNNs) have drawn increasing attention in recent years and have achieved break-throughs in graph mining tasks~\cite{kipf2016semi,velivckovic2017graph,kipf2016variational}. The input of a GNN is usually a graph containing node attributes, and after multiple layers of message passing, the model converts the input into node-level or graph-level representation, which can be used for downstream tasks such as node classification, link prediction, and graph classification.

The existing GNN models are often designed for specific tasks and need to be trained from scratch if used for any new task and new data, which will consume extra time and resources. Referring to the success of pre-training in the field of NLP~\cite{devlin2018bert} and CV~\cite{chen2020simple}, it is believed that pre-training on graph data is also beneficial. Some starting methods have verified the advantages of graph pre-training in the contexts of transfer learning~\cite{hu2019strategies}, contrastive learning~\cite{qiu2020gcc}  and graph generation~\cite{hu2020gpt}. Most pre-training methods are self-supervised, which can work on large-scale networks regardless of whether the nodes are labeled or not.

However, the existing graph pre-training methods mainly focus on capturing the structural features at node level or graph level without taking the temporal information of edge generation into consideration. The latter actually reveals the  evolution of the graph and helps to learn node representations more accurately. Therefore, the incorporation of graph dynamics into pre-training GNN models is a topic worthy of further study. 

This paper proposes a pre-training framework on dynamic graph neural networks (PT-DGNN), including two steps: firstly, sampling subgraphs in a time-aware method to generate the input for pre-training models; secondly, pre-training the GNN model on two tasks, i.e. dynamic edge generation task and node attribute generation task. The pre-training data is chronologically constrained, that is, observed edges are used to predict unobserved edges, thus capturing the evolutionary characteristics of the network.

The main contributions of this paper are as follows.

\begin{itemize}

\item In order to deal with large-scale dynamic networks, a dynamic subgraph sampling method is proposed, which can preserve the local structure in a time-aware way by sampling more newly updated nodes; 

\item a dynamic pre-training method is proposed based on graph generation tasks, which can capture structural features and node attributes by fusing temporal information;
 
\item The evaluation after fine-tuning on the link prediction task shows that our method achieves the best prediction performance in different dynamic networks.

\end{itemize}

\section{Related Work}

This section reviews related work on GNN pre-training, graph sampling and dynamic graph representation learning.

\subsection{GNN Pre-training}

Pre-training a GNN is to learn the high-quality parameters of the model at the node level and the graph level according to different tasks, and these parameters can be effectively transferred to downstream tasks. 

Hu W. et al.~\cite{hu2019strategies} proposed a pre-training method that combines  node-level and graph-level representations to train the model. At the node level, two self-supervised methods are used, namely context prediction and attribute prediction. At the graph level, supervised graph-level property prediction and structural similarity prediction are used. Combining four tasks enables GNN to learn useful local and global graph representations, and finally effectively improves the performance of graph classification tasks on molecular graphs. Hu Z. et al.~\cite{hu2019pre}  proposed a general pre-training framework that focuses on multiple subtasks, including  link reconstruction denoising, centrality score ranking and cluster preserving. The model is expected to capture the characteristics of the graph from multiple perspectives (nodes, edges, and subgraph structure). Later, they~\cite{hu2020gpt}  proposed another framework GPT-GNN, where pre-training is performed on a large-scale single graph using a graph generation task to learn node-level structural and semantic information. Specifically, subgraphs are first sampled and then sent to a general GNN model for graph generative pre-training. Qiu et al.~\cite{qiu2020gcc} proposed a self-supervised GNN pre-training framework GCC to capture the universal topological properties across multiple networks. The model designs the pre-training task as subgraph-level instance discrimination in a graph or across graphs, using contrastive learning to better learn the internal and transferable structural representation. The model is finally evaluated in three graph learning tasks (node classification, graph classification, and similarity search).

The above methods design different pre-training tasks to learn the general structural characteristics of graphs, but our work is more concerned with capturing the evolutionary characteristics of graphs by improving the pre-training tasks.

\subsection{Graph Sampling}

Graph sampling is an indispensable strategy for accelerating the training of GNN, including different types such as node-wise sampling, layer-wise sampling and subgraph sampling.

GraphSAGE~\cite{hamilton2017inductive} uses a node-wise sampling method, which aggregates a fixed number of neighbors to update node embedding. However, the common neighbors of nodes may be frequently calculated, thus consuming extra memory and time. VR-GCN~\cite{chen2018Stochastic} improves GraphSAGE by using variance reduction, which can sample fewer neighbor nodes without losing the performance. FastGCN~\cite{chen2018FastGCN} is a layer-wise sampling method, calculating the probability of neighbors being sampled based on the degree of nodes and then sampling a fixed number of nodes in each layer. The sampled subgraph is constructed as an adjacency matrix for optimization. Huang et al.~\cite{huang2018Adaptive} proposed an adaptive layer-wise sampling method, in which the neighbors sampled are shared by different parent nodes to avoid excessive expansion. Moreover, the method is adaptive and can reduce the sampling variance explicitly. Cluster-GCN~\cite{chiang2019Cluster} uses graph clustering algorithm to cut the whole graph into several small clusters, randomly select $q$ clusters to form subgraphs, and then train on the sampled subgraphs. Similar to Cluster-GCN, GraphSAINT~\cite{chen2020Graphsaint} also samples the subgraphs from the original graph as the input of GCN. The difference is that the latter explicitly minimize the variance caused by the sampling, which can ensure that node aggregation after sampling is unbiased. LADIES~\cite{zou2019layer} considers subgraph sampling from two perspective: layer-wise and node-wise. The model constructs a bipartite graph between sampled nodes and their neighbors and calculates the sampling probability according to the degree of the candidate nodes. The sampling is completed by repeatedly updating the candidate pool. It is verified that LADIES can reduce the sampling variance and maintain the denseness of the subgraph.

\subsection{Dynamic Graph Representation Learning}

Graph representation learning (also known as graph embedding) aims to learn effective low-dimensional vector representations of nodes, edges, or graphs, which can be used for downstream tasks of classification or prediction. The initial methods are based on network proximity~\cite{tang2015line} or statistics derived from random walks~\cite{perozzi2014deepwalk,grover2016node2vec}. But they rely on specific networks and have poor adaptability. After GNN ~\cite{Scarselli2009Graph} was proposed, a series of GNN models emerged, such as GCN~\cite{kipf2016semi}, GAT~\cite{velivckovic2017graph}, GAE~\cite{kipf2016variational}, GraphSAGE~\cite{hamilton2017inductive} and GraphInfomax~\cite{velickovic2019deep}.

Note that most graphs are constantly changing over time (especially in edge generation), a series of methods are proposed to study the influence of dynamics on graph embedding. In DANE~\cite{li2017attributed}, the graph representation at the time $t$ is updated based on the variation of the adjacency matrix and the attribute matrix over time. The parameters to initialize model at current time step are directly derived from the previous time step, thus speeding up the training. DynGEM~\cite{goyal2018dyngem} splits the dynamic graph into different snapshots, and uses SDNE (Structural Deep Network Embedding)~\cite{wang2016structural} method for each snapshot. HTNE~\cite{zuo2018embedding}  introduces Hawkes process theory into the dynamic node representation model, by considering that the impact of neighbors on the central nodes will weaken over time. DyRep~\cite{trivedi2018representation} deals with two dynamic processes: association process and communication process. The former focuses on the change of topology, while the latter deals with the change of node interactions. DySAT~\cite{sankar2018dynamic} combines attention~\cite{vaswani2017attention}  and GNN to learn dynamic graph representation from two aspects: structural attention and temporal attention, both using the past representation of nodes to update the current representation. Unlike the above methods, CTDNE~\cite{nguyen2018continuous} proposes a temporal random walk sequence and then uses skip-gram to get node representation. The method effectively solves the problem of time information loss in the snapshot-based methods. TGN~\cite{Rossi2020Temporal} learns node representation by using a memory module, multiple message function and graph-based operators, effectively solving the problem of time information loss in snapshot-based methods. The above works have verified that the quality of node representation can been improved after learning from graph dynamics.

\section{Pre-training on Dynamic GNN}

This section details the proposed pre-training method based on dynamic attributed graph generation.

\subsection{Problem Description}

\begin{definition}[Dynamic Networks]
In a dynamic network, the interaction between nodes occurs at a specific time, and the network structure changes over time. In this paper, the dynamic network is formalized as a graph $G = (V,\xi_t,\tau,X)$, where $V$ represents the node set, $\xi_t \subseteq V \times V\times R^+$ represents the edge set with timestamp $t(t \in R^+)$, and $\tau:\xi \rightarrow R^+$ is a function that maps each edge to a corresponding timestamp t. Any given edge $e=(u,v,t) \in \xi_t$  represents the interaction between node $u$ and $v$ at time $t=\tau(u,v)$. $X$ denotes the matrix of node attributes.
\end{definition}

\begin{problem}[Dynamic Network Representation Learning]
For a dynamic network $G = (V,\xi_t,\tau,X)$, the task of dynamic network representation learning is to learn a mapping function $f_\theta:V\rightarrow{R}^d $, which outputs the node features in a low-dimensional space. The node representation can embed  multi-type of information, such as structure, semantics (node attributes) and evolution (time sequence of edges), etc.
\end{problem}

\begin{problem}[Pre-Training on Dynamic GNN]
In this paper, pre-training on dynamic GNN refers to the use of graph generation tasks that take into account the edge timestamps, to learn general features (including evolutionary information) from dynamic graphs. After pre-training, the parameter $\theta$ of the model $f_\theta$ is obtained.
\end{problem}

\subsection{The PT-DGNN Framework}

\begin{figure*}[htbp]
	\centering
	\includegraphics[scale=0.31]{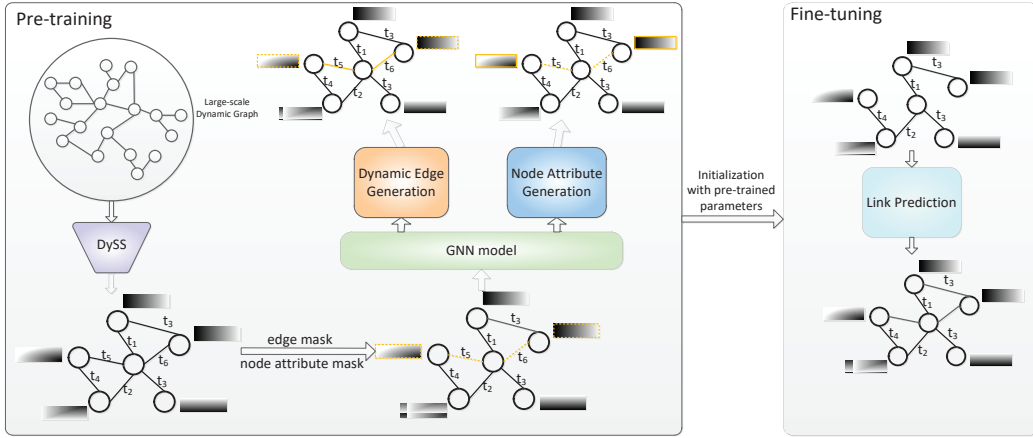}
	\caption{Overall framework of PT-DGNN}
	\label{frame}
\end{figure*}

This paper proposes a pre-training framework, named PT-DGNN (Figure \ref{frame}), which includes dynamic subgraph sampling (DySS), dynamic attributed graph generation and fine-tuning.

The input is a large-scale dynamic graph $G = (V,\xi_t,\tau,X)$. After pre-training,  a general GNN model $f_\theta$ is learned and can be fine-tuned in a specific task such as link prediction.

\subsection{Dynamic Subgraph Sampling}

When pre-training a GNN model on large-scale graphs, subgraph sampling is usually required~\cite{zou2019layer}. In this paper, a dynamic subgraph sampling (DySS) method is proposed to generate the input required for pre-training.

\begin{algorithm}[h]
	\small
	\caption{DySS}
	\label{Dyss}
	\textbf{Input}: dynamic graph $DG$, sampling depth $d$, sampling width $w$ \\
	\textbf{Output}: dynamic subgraph $SG$ 
	\begin{algorithmic}[1] 
		\STATE Randomly select $w$ starting nodes to $SN$;
		\STATE Create a candidate pool $CPool$;
		\STATE $PS$ $\leftarrow$ $\{0\}$; $//$sampling probability of all nodes
		\STATE $SG$ $\leftarrow$ $SN$;
		\FOR {$i=1$ to $d$}
		\FOR{each $v$ in $SN$}
		\FOR{each $u$ in Neighbor($v$)}
		\STATE $CPool$ $\leftarrow$ $CPool$ $\cup$ $\{u\}$;
		\STATE $PS_u$ $\leftarrow$ $PS_u$ + $\tau(v,u)$;
		\ENDFOR
		\ENDFOR
		\STATE Update $PS$ based on Eq.\ref{prob} ;
		\STATE Select $\{v_{1}^{'},...,v_{w}^{'}\}$ according to $PS$;
		\STATE $SG$ $\leftarrow$  $SG$ $\cup$ $\{v_{1}^{'},...,v_{w}^{'}\}$;
		\STATE $CPool$ $\leftarrow$ $CPool$ $-$ $\{v_{1}^{'},...,v_{w}^{'}\}$;
		\STATE $SN$ $\leftarrow$ $SN$ $\cup$ $\{v_{1}^{'},...,v_{w}^{'}\}$;
		\ENDFOR
		\STATE \textbf{return} $SG$
	\end{algorithmic}
\end{algorithm}

The DySS method is described in Algorithm \ref{Dyss}, which mainly includes four steps: 1) Randomly select $w$ nodes as the  initial sampling point; 2) Put the first-order neighbors of each initial sampling node into the candidate pool and set their sampling probability(line 6-11); 3) Select final $w$ nodes from the candidate pool as the sampled nodes according to their updated sampling probability calculated by Eq. \ref{prob} and update the initial sampling point (line 12-16); 4) Repeat step 2 and 3 for $d$ times to obtain the subgraph.  Here, $d$ and $w$ are two hyper-parameters, which represent the sampling depth and width respectively.



The sampling probability of nodes in the candidate pool ($CPool$) is normalized by Eq. \ref{prob}

\begin{align}
	\label{prob}
	\widetilde{PS}_{u} = \frac{PS_{u}^2}{\sum_{k\ \in\ {CPool}} PS_{k}^2}  \quad {{\forall u} \in CPool}
\end{align}

Compared to LADIES~\cite{zou2019layer} which is designed for static graphs, DySS is able to sample time-aware subgraphs. 

\subsection{Dynamic Attributed Graph Generation}

In order to learn network dynamics, we use two pre-training tasks, i.e. dynamic edge generation and node attribute generation, and each node in the network must participate in these two tasks. In dynamic edge generation, a time-based mask method is used to predict node $i$'s unobserved edges $\xi_{i,f}$  with node $i$'s observed edges $\xi_{i,h}$ and node $i$'s attributes $X_i$. In node attribute generation, the attributes of node $i$'s neighbors are used to predict node $i$'s attributes $X_i$.

The autoregressive generation process can be defined in Eq. \ref{reg}.  For any node $i$,  $X_{<i}$ is the attribute set of the nodes that are generated before $i$ and $\xi_{<i}$ is the set of all edges in the graph before $i$ is generated. 
$X_i$ is the attribute of $i$ and  $\xi_i$ is the set of edges connected to $i$. $p_\theta$ indicates the likelihood that uses $X_{<i}$ and $\xi_{<i}$ to generate $X_i$ and  $\xi_i$.

\begin{align}
	\label{reg}
	&p_\theta\left(X_i,\xi_i\middle|X_{<i},\xi_{<i}\right)\\
	&=\sum_{h}{p_\theta\left(X_i,\xi_{i,f}\middle|\xi_{i,h},X_{<i},\xi_{<i}\right) \cdot p_\theta\left(\xi_{i,h}\middle|X_{<i},\xi_{<i}\right)}\nonumber\\
	&=\mathbbm{E}_h\left[p_\theta\left(X_i,\xi_{i,f}\middle|\xi_{i,h},X_{<i},\xi_{<i}\right)\right]\nonumber,\end{align}

Assuming that the generation of  $X_i$ and  $\xi_{i,f}$ are independent, $p_\theta\left(X_i,\xi_{i,f}\middle|\xi_{i,h},X_{<i},\xi_{<i}\right)$ can be simplified to Eq. \ref{sim}.
\begin{align}
	\label{sim}
	&\left[p_\theta\left(X_i,\xi_{i,f}\middle|\xi_{i,h},X_{<i},\xi_{<i}\right)\right]\\
	&=\left[p_\theta\left(X_i\middle|\xi_{i,h},X_{<i},\xi_{<i}\right) \cdot p_\theta\left(\xi_{i,f}\middle|{\xi_{i,h},X}_{\le i},\xi_{\le i}\right)\right],\nonumber 
\end{align}
where the first term $p_\theta\left(X_i\middle|\xi_{i,h},X_{<i},\xi_{<i}\right)$ denotes the generation of attributes for node $i$,  and the second term  $p_\theta\left(\xi_{i,f}\middle|{\xi_{i,h},X}_{\le i},\xi_{\le i}\right)$ refers to the prediction of unobserved edges for node $i$.

\subsubsection{Dynamic Edge Generation}

The time sequence of edge generation implies the evolutionary law of the network. The generation of late interactions is often related to that of early interactions (See Example \ref{exp1}).

\begin{figure}[htbp]
	\centering
	\includegraphics[width=14cm]{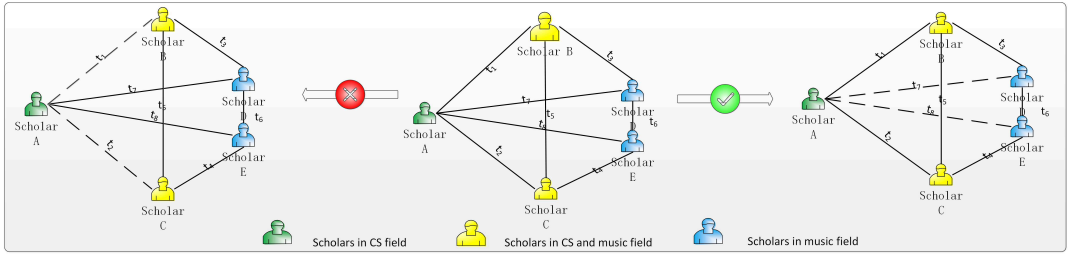}
	\caption{An illustrative example of time-based mask method}
	\label{example}
\end{figure}

\begin{example}[Dynamic Edge Generation Example]
	\label{exp1}
As shown in Figure \ref{example}, taking the collaboration network as an example, suppose that A is a scholar in the computer science field, B and C are scholars in the cross-field of computer science and music, D and E are scholars in the music field. Generally, A is unlikely to cooperate with scholars D and E in the music field unless he/she cooperates with B and C first. Therefore, the order of edge generation in the local graph is more likely to be $\{(A,B),(A,C)\} \rightarrow \{(A,D),(A,E)\}$ than $\{(A,D),(A,E)\} \rightarrow \{(A,B),(A,C)\}$.
\end{example}

Instead of randomly masking edges in previous methods, we propose a time-sensitive edge mask method, which follows the assumption that historical neighbor information should be used to predict new neighbors, not vice versa. The method is described in Algorithm \ref{mask_edge}. For any node $v$, the edges between itself and its neighbors, if having newer timestamp, should be more probably masked.

\begin{algorithm}[h]
	\small
	\caption{Time-based edge mask}
	\label{mask_edge}
	\textbf{Input}: subgraph $SG$, number of masked edges $N$\\
	\textbf{Output}: set of masked edges  $ME$
	\begin{algorithmic}[1] 
		\STATE $PT \leftarrow \{0\}$; $//$ mask probability of all nodes.
		\FOR{each $v$ in $SG$}
		\FOR{each $u$ in Neighbor($v$)}
		\STATE $PT_u$ $\leftarrow$ $PT_u$ + $\tau(v,u)$;
		\ENDFOR
		\STATE $PT$ $\leftarrow$ $Softmax(PT)$;
		\STATE Select $\{u_1,...,u_N\}$ according to $PT$;
		\STATE $ME$ $\leftarrow$ $ME \cup \{u_1,...,u_N\}$;
		\ENDFOR
		\STATE \textbf{return} $ME$
	\end{algorithmic}
\end{algorithm}

The GNN model uses the masked graph as input and generates node embedding $r^E$ by learning from dynamic edge generation. To optimize the model, the contrastive loss $L^E$ is defined as:

\begin{equation}
	\label{edge}
	L^{E}=-\sum_{i\ \in\ V}\sum_{\ \ \quad\mathclap{j^+\in\ \xi_{i,f}}\ }{log\frac{exp\left(Sim\left(r_i^{E},r_{j^+}^{E}\right)\right)}{\sum_{j\in\ \xi_{i,f} \cup {S_i}} exp\left(Sim\left(r_i^{E},r_j^{E}\right)\right)}} ,
\end{equation}
where $S_i$ is the set of unconnected node pairs where one node is $i$, the  $Sim\left(r_i^E, r_j^E\right)$, $\forall i,j\in V $   function is used to calculate the embedding similarity of node $i$ and node $j$. In this paper, the function is defined as $r_i^E \cdot r_j^E$.

By optimizing $L^E$, the generation probability of unobserved edges is maximized, so that the pre-training process learns the structural evolution of the graph.

\subsubsection{Node Attribute Generation}

In the node attribute generation task, node attributes are masked and replaced with a learnable vector $X^{'}$. The GNN model uses the masked graph to generate node embedding $r^A$. Here, a multi-layer perceptron is used as a decoder which takes $r^A$ as input and generates the masked attributes. To optimize the model, $L^A$ is defined, which is L2 distance function to measure similarity between generated attributes and masked attributes.

\begin{equation}
	\label{attr}
	L^{A}=\sum_{i\ \in\ V}\left|\left.MLP\left(r_i^{A}\right)-X_i\right|\right.^2	
\end{equation}

By optimizing $L^A$, the generation probability of node attributes is maximized, so that the pre-training process learns the semantic information of the graph.

Even though the nodes have two kinds of embeddings $r^E$ and $r^A$ in this pre-trained model, only  $r^E$ is used to conduct GNN information propagation between GNN layers as the real node attributes are masked in $r^A$.

\subsection{Fine-tuning in Link Prediction}
In the fine-tuning stage, $f_\theta$ learned by pre-training is used to initialize the GNN model and the input to train the model is a new graph. For link prediction task, a loss function $L^F$ based on dynamic graph is defined.

\begin{equation}
	L^{F}=\sum_{i\ \in\ V}-log\left(\sigma\left(r_i^Tr_j\right)\right)-Q\cdot {E_{j_n \sim P_n\left(j\right)}}log\left(\sigma\left(r_i^Tr_{r_n}\right)\right),
\end{equation}
where $j$ is a node that co-occurs near $i$ on a fixed-length temporal random walk~\cite{nguyen2018continuous}, $\sigma$ is sigmoid function, $P_n$ is the negative sampling distribution, and $Q$ defines the number of negative samples. 

By optimization, the connected nodes are closer in the latent space while unconnected nodes can be more distinguishable.

\section{Experiments}

Comparative experiments are conducted on three dynamic network datasets and the link prediction task is used to evaluate the performance of different models.

\subsection{Datasets}

Three realistic dynamic network datasets (HepPh, MathOverflow and SuperUser) were selected from the SNAP website\footnote{\url{http://snap.stanford.edu/}}. \textit{HepPh} is a citation network dataset in high energy physics phenomenology constructed from the e-print arXiv. It covers papers from January 1993 to April 2003. If paper $i$ cited paper $j$ in a certain time $t$, there is a directed edge with time $t$ from node $i$ to node $j$. \textit{MathOverflow} is a temporal network of interactions on the stack exchange website Math Overflow. A directed edge $(u,v,t)$ indicates that user $u$ and user $v$ interacted at time $t$. The interactions include answering questions, commenting question and commenting answer. \textit{SuperUser}  is another temporal network of interactions on the stack exchange web site Super User. It has the same interaction types with the MathOverflow dataset, but on a much larger scale.

The statistics of three datasets are shown in Table \ref{datasets}.

\begin{table}[htbp]
	\centering
	\resizebox{8cm}{!}{
	\begin{tabular}{l|rrrr}
		\toprule
		Datasets	& Nodes & Edges&\makecell[r]{Time span\\ (days)} & Types\\
		\midrule
		HepPh       & 34,546  	& 421,578 	&3,720 &Directed \\
		MathOverflow &  24,818 & 239,978  	&2,350 &Directed \\
		SuperUser	  & 194,095 & 924,886 &2,773&Directed \\
		\bottomrule
	\end{tabular}
	}
	\caption{Statistics of the datasets}
	\label{datasets}
\end{table}

\subsection{Comparison Methods}

In the experiment, related models are compared, including static GNN models which can potentially be used to pre-train GNNs~\cite{hu2020gpt}, a static GNN with pre-training and a dynamic GNN with pre-training (i.e. ours).

\paragraph{node2vec\rm{~\cite{grover2016node2vec}}} This method uses biased random walk sequences as the input of Skip-Gram to learn the representation of nodes in the graph. It is a benchmark method for comparison.
\paragraph{GAE and VGAE\rm{~\cite{kipf2016variational}}} Both are graph auto-encoders. The former uses GCN as the encoder, and the latter uses GCN encoder to learn the distribution of hidden node representations. Both use the inner product as the decoder to reconstruct the graph.
\paragraph{GraphSAGE\rm{~\cite{hamilton2017inductive}}} In this model, both first-order and second-order neighbor information of nodes is used during graph convolution. The optimization goal is to make neighboring nodes have more similar representations.
\paragraph{GraphInfomax\rm{~\cite{velickovic2019deep}}} This model combines the mutual information theory and tries to maximize the local-global mutual information between node embedding and graph embedding. 
\paragraph{CTDNE\rm{~\cite{nguyen2018continuous}}} It is a dynamic network embedding model also based on random walk and Skip-Gram. But the  model uses the temporal random walk, that is, the edge sequence is in chronological order.
\paragraph{HTNE\rm{~\cite{zuo2018embedding}}} It is also a dynamic network embedding model, introducing the Hawkes process theory. It is based on the fact that the influence of neighbors on the central nodes decays over time.
\paragraph{GPT-GNN\rm{~\cite{hu2020gpt}}} It is a pre-training model for static networks, which also uses edge generation task and node attribute generation task to pre-train a single graph. In GPT-GNN, a randomly mask method is used in edge generation task.

\subsection{Experimental Setup}

According to the time sequence of the edges, we split each dataset into four parts, which are used for pre-training, training, verification and testing.  The data ratio of each part is 7:1:1:1. For models without pre-training, the first two parts of the data are merged as the training set.

Due to the lack of node attributes in public dynamic network datasets, we use node2vec to generate virtual attributes of nodes over the entire time span. The initialization method is commonly used in dynamic network representation learning methods~\cite{singer2019node,gong2020exploring}.

For all models, the hidden dimension is set to 400 and the number of GNN layer is set to 3. The pre-training hyper-parameters of PT-DGNN remain the same as GPT-GNN. The depth and width of the sub-graph sampling are set to 6 and 128, respectively. The ratio of mask edges is 50\%. For comparison, the fine-tuning task of GPT-GNN is also set to link prediction instead of node classification. Both pre-training models are optimized via AdamW, and 20 epochs are conducted for both pre-training and fine-tuning.

The PT-DGNN is implemented by PyTorch Geometric package\footnote{\url{https://pytorch-geometric.readthedocs.io/en/latest/#}}, and the source code is available on the website\footnote{\url{https://github.com/Mobzhang/PT-DGNN/}}.

\subsection{Experimental Results}

\paragraph{Comparison of Performance on Link Prediction}
The link prediction task is widely used to evaluate the performance of dynamic graph embedding~\cite{nguyen2018continuous, gong2020exploring}. Each method was run ten times on each dataset and the average AUC, AP and F1-score are reported in Table 2-4. The results of the CTDNE method are missing on the SuperUser dataset due to its difficulty to train large-scale dynamic networks. PT-DGNN$_{Attr}$ and PT-DGNN$_{Edge}$ are variants of PT-DGNN, using one of the two pre-training tasks. DGNN$_{npt}$ is a simplified version without pre-training, which is directly trained on the sampled subgraphs.

	\begin{table}[htpb]
		\centering
		\small
		\begin{tabular}{|l|c|c|c|}\hline
		    \multirow{2}{*}{Methods}
		    &\multicolumn{3}{c|}{HepPh}  \\  \cline{2-4}		    
			&AUC&AP&F1 \\ \cline{1-4}			
			Node2vec &	0.6235±3.9\%&	0.6192±4.2\%&	0.5218±2.9\%	 \\ 
			GAE&	0.7134±4.5\%&	0.7027±1.8\%&	0.6224±3.4\%\\ 
			VGAE&	0.7189±4.1\%&	0.7102±3.1\%&	0.6392±4.5\%\\ 
			GraphSAGE&	0.5998±5.2\%&	0.5743±5.1\%&	0.5149±6.8\%  \\ 
			GraphInfomax&	0.6203±7.6\%&	0.5844±8.3\%&	0.5246±9.7\% \\ 
			CTDNE&	0.7429±2.27\%&	0.6628±2.89\%&	0.7103±1.45\%	\\ 
			HTNE&	0.6948±3.16\%&	0.6793±1.63\%&	 0.6631±4.25\% \\ 
			GPT-GNN&	0.8546±2.51\%&	0.8278±3.54\%&	0.7803±1.7\% \\ 
			DGNN$_{npt}$&	0.8586±1.9\%&	0.8302±3.32\%&	0.7785±3.37\% \\
			\textbf{PT-DGNN$_{Attr}$}&	\textbf{0.8617}±2.33\%&	\textbf{0.8295}±2.97\%&	\textbf{0.7807}±1.69\% \\
			\textbf{PT-DGNN$_{Edge}$}&	\textbf{0.8684}±1.22\%& 	\textbf{0.8314}±1.55\%&	\textbf{0.7834}±1.49\% \\
    			\textbf{PT-DGNN(ours)}&	\textbf{0.8762}±0.91\%&	\textbf{0.8327}±1.55\%&	\textbf{0.7865}±1.26\%	 \\\cline{1-4}
		\end{tabular}
		\caption{Comparative results in link prediction on HepPh dataset}
	\end{table}
	
		\begin{table}[htpb]
		\centering
		\small
		\begin{tabular}{|l|c|c|c|}\hline
		    \multirow{2}{*}{Methods}
		    &\multicolumn{3}{c|}{MathOverflow}  \\  \cline{2-4}		    
			&AUC&AP&F1 \\ \cline{1-4}			
			Node2vec &	0.6250±4.1\%&	0.62158±3.6\%& 0.5324±4.1\%	 \\ 
			GAE&	0.6879±3.1\%&	0.6844±1.3\%&	0.5165±5.4\%\\
			VGAE&	0.6935±2.5\%&	0.6997±0.9\%& 0.5568±3.4\%\\ 
			GraphSAGE&	0.6696±4.8\%&	0.6619±4.8\%&	0.5571±5.7\%  \\
			GraphInfomax&	0.6730±6.6\%&	0.6753±5.4\%&	0.5743±6.5\% \\
			CTDNE&	0.7432±0.64\%&	0.7916±0.39\%&	0.6543±0.81\%	\\
			HTNE&	0.7493±0.39\%& 0.6924±0.52\%&	 0.6414±0.51\% \\
			GPT-GNN &	0.8910±0.98\%&	0.9023±1.05\%& 0.8079±1.96\% \\
			DGNN$_{npt}$ &	0.8963±1.35\%&	0.8823±2.23\%&	0.8061±2.4\% \\
			\textbf{PT-DGNN$_{Attr}$}&	\textbf{0.9037}±1.08\%&	\textbf{0.9017}±1.39\%& \textbf{0.8094}±1.52\% \\
			\textbf{PT-DGNN$_{Edge}$}&	\textbf{0.9064}±1.24\%& 	\textbf{0.9035}±1.31\%&	\textbf{0.8123}±2.18\% \\
			\textbf{PT-DGNN(ours)}& \textbf{0.9196}±1.07\%&	\textbf{0.9044}±1.5\%&	\textbf{0.8154}±2.46\%	 \\\cline{1-4}		
		\end{tabular}
		\caption{Comparative results in link prediction on MathOverflow dataset}
	\end{table}
	
		\begin{table}[htpb]
		\centering
		\small
		\begin{tabular}{|l|c|c|c|}\hline
		    \multirow{2}{*}{Methods}
		    &\multicolumn{3}{c|}{SuperUser}  \\  \cline{2-4}		
			&AUC&AP&F1 \\ \cline{1-4}		
			Node2vec &	0.5841±5.1\%&	0.5961±3.4\%& 0.5147±3.7\%	 \\ 
			GAE&	0.6985±3.8\%&	0.6803±1.8\%&	0.5271±3.2\%\\ 
			VGAE&	0.7064±3.6\%&	0.6894±2.7\%& 0.5431±2.9\%\\ 
			GraphSAGE&	0.6582±4.1\%&	0.6693±6.1\%&	0.5618±5.6\%  \\ 
			GraphInfomax&	0.6849±7.8\%&	0.6827±7.4\%&	0.5820±8.5\% \\ 
			HTNE&	0.6965±0.47\%& 0.6536±0.39\%&	 0.6101±0.69\% \\ 
			GPT-GNN&	0.8455±2.74\%&	0.8568±2.8\%& 0.7462±2.57\% \\
			DGNN$_{npt}$&	0.8524±2.58\%&	0.8526±4.15\%&	0.7538±3.01\% \\
			\textbf{PT-DGNN$_{Attr}$}&	\textbf{0.8671}±3.41\%&	\textbf{0.8541}±3.97\%&\textbf{ 0.7549}±3.46\% \\
			\textbf{PT-DGNN$_{Edge}$}&	\textbf{0.8642}±3.54\%& 	\textbf{0.8588}±4.41\%&	\textbf{0.7581}±3.48\% \\
			\textbf{PT-DGNN(ours)}& \textbf{0.8750}±3.38\%&	\textbf{0.8626}±4.19\%&	\textbf{0.7559}±4.7\%	 \\\cline{1-4}
		\end{tabular}
		\caption{Comparative results in link prediction on SuperUser dataset}
	\end{table}
	
The experimental results show that the average performance of the models after pre-training on sampled graphs is significantly higher than those without pre-training. When comparing models without pre-training, the performance of CTDNE and HTNE is slightly higher than static GNN models, but their performance is not stable on different datasets. Since both the training and evaluation of DGNN$_{npt}$ use subgraphs obtained through dynamic sampling,  its performance is much better than the method using the full graph for training and evaluation, and is almost equivalent to the GPT-GNN method. PT-DGNN performs better than the pre-trained GNN model GPT-GNN. For example, the AUC value on the three datasets have increased by 2\% to 3\%. Moreover, the performance of PT-DGNN using a single pre-training task (edge generation or attribute generation) is also better than or comparable to that of the GPT-GNN method, which pre-trains on both tasks. It fully verified the effectiveness of dynamic subgraph sampling and the time-based mask method.

\subsubsection{Comparison of different GNN base models}

In order to further explore the performance of the PT-DGNN framework, we use different GNN base models including GCN~\cite{kipf2016semi}, GAT~\cite{velivckovic2017graph}, ARMA~\cite{2021Graph} and SGC~\cite{2019Simplifying} to compare PT-DGNN and DGNN$_{npt}$ on the HepPh dataset. 

\begin{figure}[htbp]
	\centering
	\includegraphics[scale=0.37]{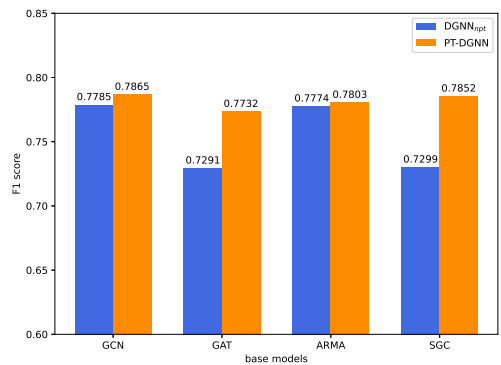}
	\caption{Performance of PT-DGNN with different base GNN models}
	\label{gnn}
\end{figure}

As shown in Figure \ref{gnn}, the F1-score of  four PT-DGNN variants are all superior to that of DGNN$_{npt}$, verifying the effectiveness of the pre-training framework. Among them, the performance improvement of PT-DGNN based on GAT and SGC is more significant. The attention mechanism in GAT can capture more accurate graph structure information, thus highlighting the advantages of the pre-training framework. SGC has fewer parameters than GCN, and the pre-training framework can provide more accurate initialization parameters, which greatly improve prediction performance.  For ARMA, its auto-regressive moving average filter can already capture the global structural information of the graph, so it may  be difficult to get a higher improvement for the pre-training framework  also focuses more on learning global information.

\subsubsection{Comparison of Subgraph Sampling Methods}

\begin{figure}[htbp]
	\centering
	\includegraphics[scale=0.37]{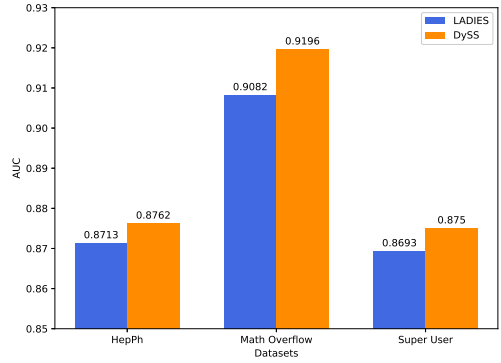}
	\caption{Performance of PT-DGNN with different sampling methods}
	\label{sample}
\end{figure}

In PT-DGNN, we also compared DySS (the sampling method in this paper) and the LADIES~\cite{zou2019layer} method in average AUC value . The latter only uses neighborhood structure to sample nodes, while ours also uses the timestamps of neighborhood edges. As shown in Figure \ref{sample}, PT-DGNN using DySS performs  better than that using LADIES in average AUC value, indicating that the use of time information helps to generate more representative subgraphs. 

\subsubsection{Impact of Subgraph Sampling Size}

\begin{figure}[h]
	\centering
	\subfigure[HepPh]{
		\label{hp}
		\includegraphics[scale=0.27]{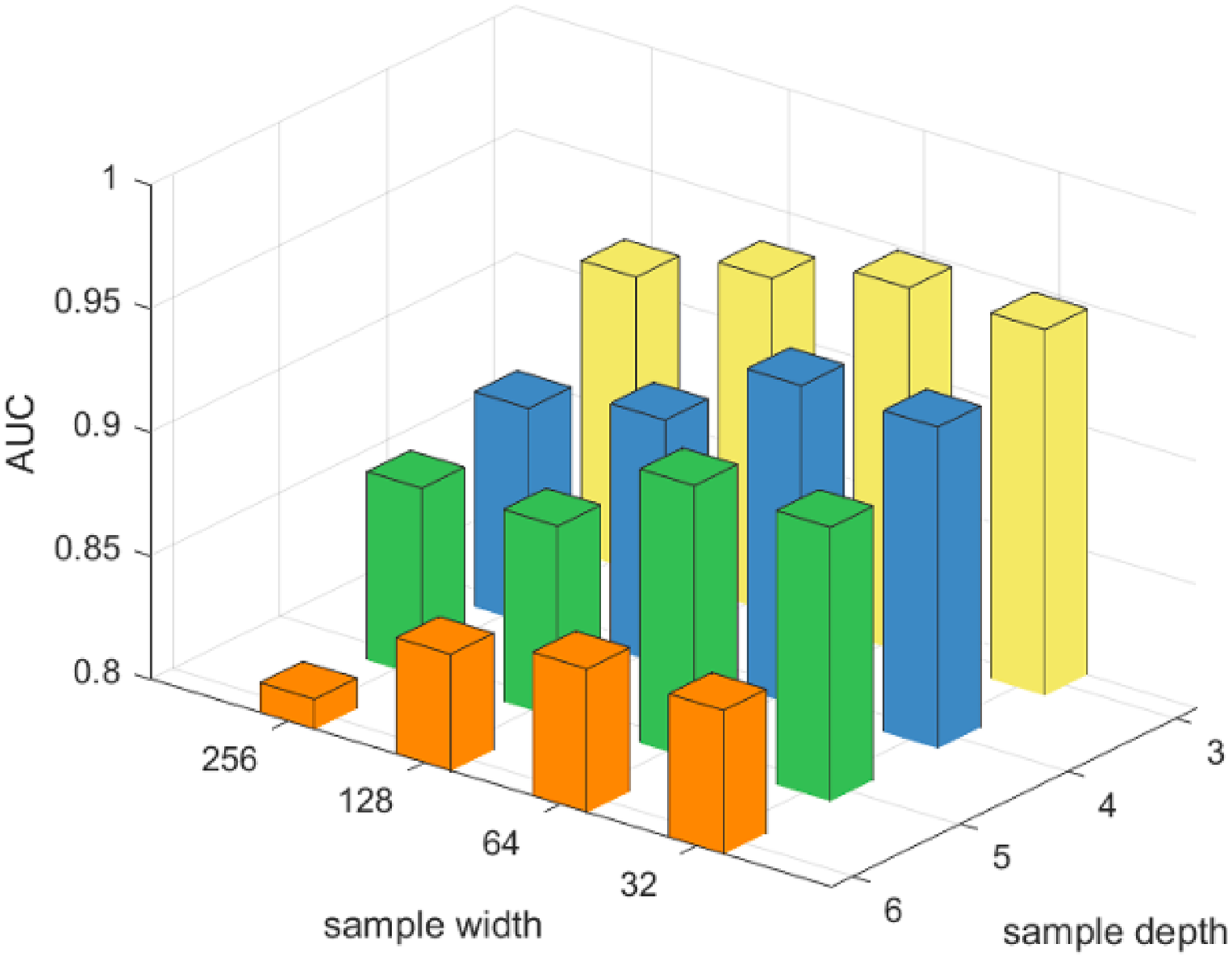}} 
	\subfigure[MathOverflow]{
		\label{mo}
		\includegraphics[scale=0.27]{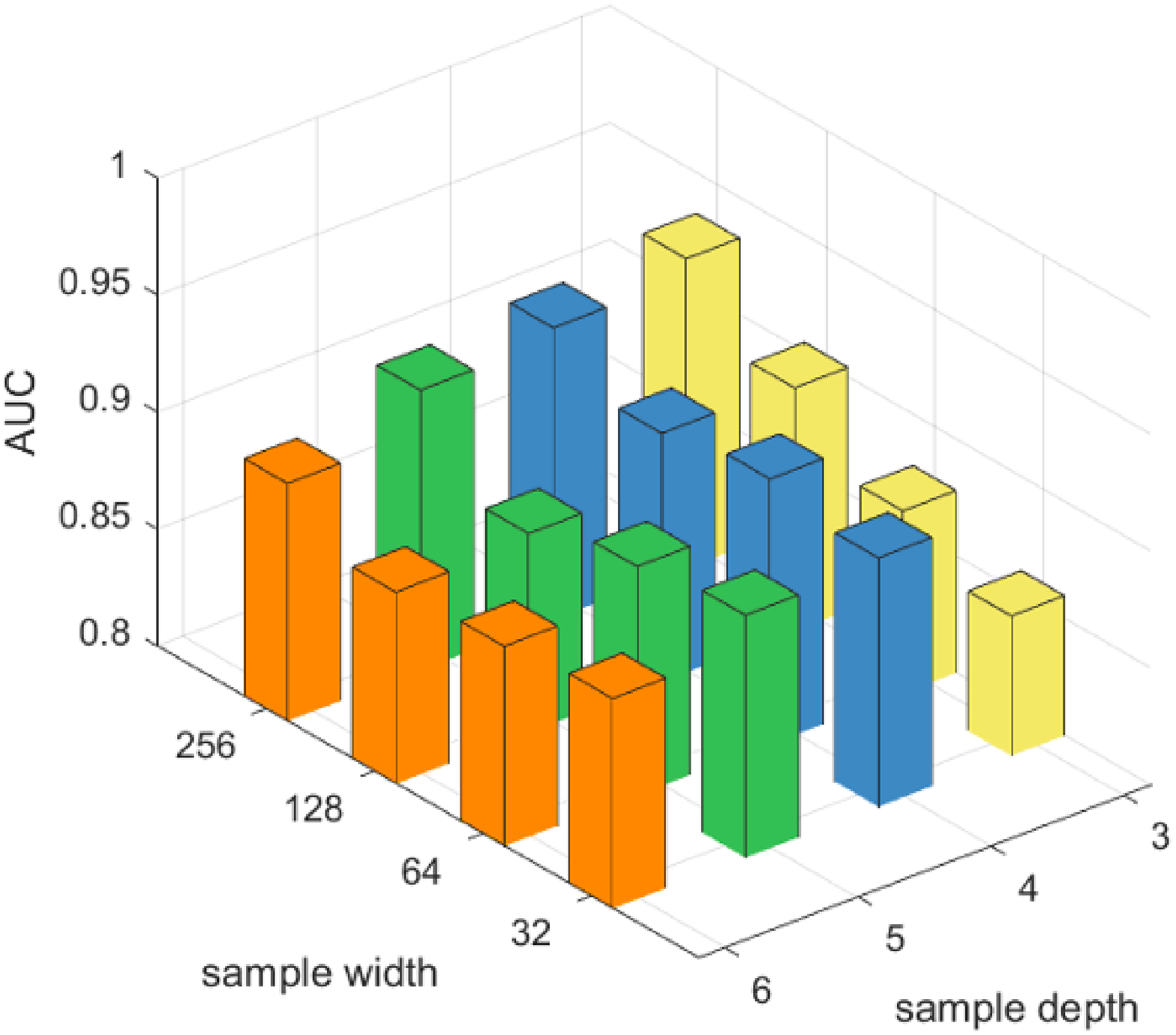}}
	\subfigure[SuperUser]{
		\label{su}
		\includegraphics[scale=0.27]{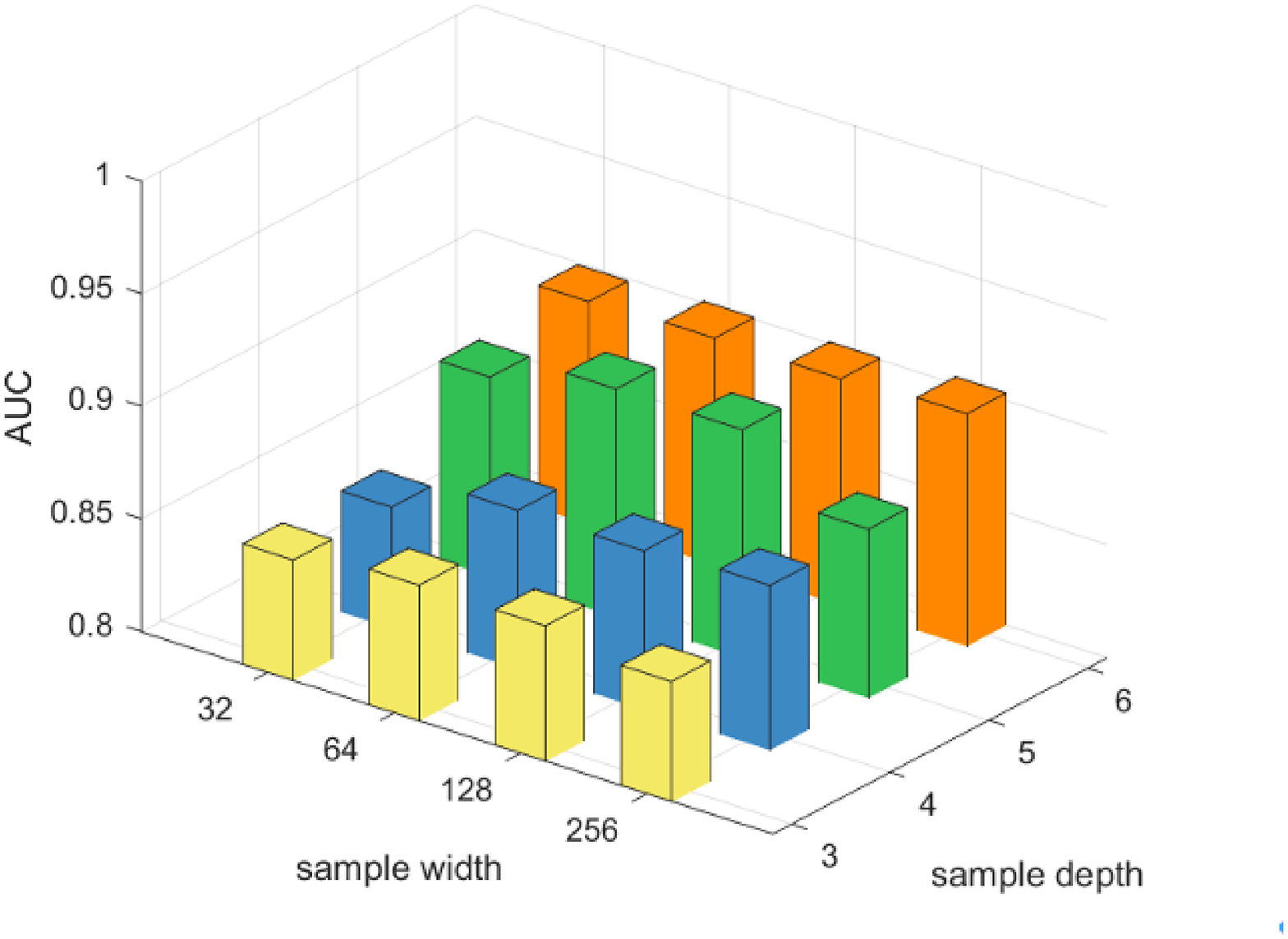}}
	\caption{The impact of sampling size on the results.}
	\label{samplesize}
\end{figure}

We analyzed the impact of the subgraph sampling size (from the depth and width dimensions) on the link prediction results. The results in Figure \ref{hp}-\ref{su} show that for the large-scale graph, the larger the size of the sampled subgraph, the better the performance can be obtained. When the scale of the original graph is not so large, the smaller the size of the sampled subgraph, the better the result. But the effect of the sampling width on the results does not follow obvious rules.

\subsubsection{Impact of Pre-training Data Size}
\begin{figure}[htbp]
	\centering
	\includegraphics[scale=0.28]{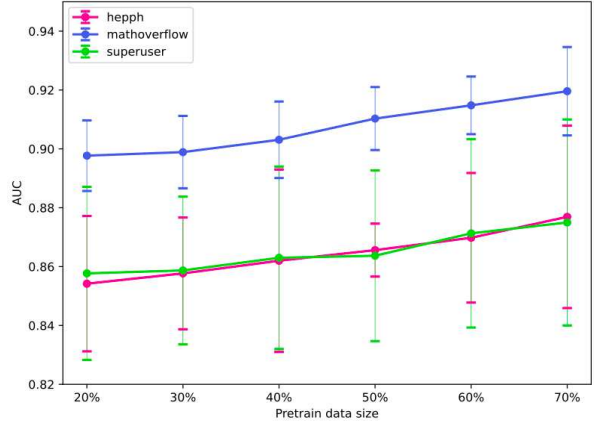}
    \caption{The impact of pre-training data size on the results.}
    \label{ptsize}
\end{figure}

We further explore the impact of the data size (the proportion of the entire dataset) of the pre-training on the final result. The performance of PT-DGNN and GPT-GNN on the three datasets are compared in Figure \ref{ptsize}. It shows that as the amount of pre-training data increases, the model can learn more robust parameters from it, so as to better serve the fine-tuning task.

\subsubsection{Impact of Mask Edge Ratio}

\begin{figure}[htbp]
	\centering
	\includegraphics[scale=0.28]{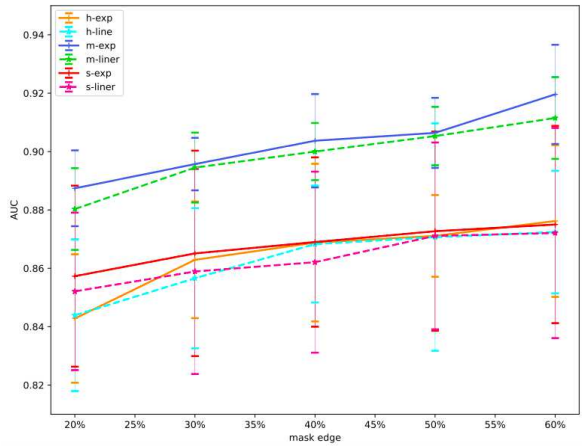} 
    \caption{The impact of ratio of masked edges on the results.}
    \label{sub2}
\end{figure}

In the dynamic edge generation module of PT-DGNN, the softmax function is used to calculate the probability of the edge being masked. In the experiment, we also used the linear function to calculate the probability for comparison. The results in Figure \ref{sub2} show that the softmax function is slightly better than the linear function in the mask strategy. As the number of masked edges increase, the performance of the PT-DGNN model gradually improves. But when the ratio continues to increase, isolated nodes may be generated making it difficult to train the model.

\section {Conclusion}

In this paper, a pre-training method for dynamic GNNs is proposed. Different from other graph-based pre-training methods, our method takes into account the dynamics of network evolution and combines the structural features with the temporal features of the edges in the graph generation task to learn more accurate low-dimensional representation

 of nodes. For this purpose, we propose a time-based edge mask method and a time-based subgraph sampling strategy. We fine-tune the pre-trained model in link prediction on multiple datasets, and the experimental results verify the superiority of our method.

Future work will mainly explore other pre-training tasks that use time information, such as time-based attribute generation tasks, and use the idea of some contrastive learning methods such as CPC (Contrasitve Predictive Coding)\cite{oord2019representation} for dynamic graph learning.

\section*{Acknowledgement}
This research was supported by the National Natural Science Foundation of China (Grant No. 61772284 and 61876091) and the Natural Science Foundation of Nanjing University of Posts and Telecommunications (Grant No. NY221071).

\bibliographystyle{elsarticle-num}

\bibliography{mybib}

\end{document}